\documentclass[10pt,twocolumn,letterpaper]{article}

\usepackage[pagenumbers]{cvpr}      
\makeatletter
\@namedef{ver@everyshi.sty}{}
\makeatother
\usepackage{tikz}
\usepackage{graphicx}
\usepackage{amsmath}
\usepackage{amssymb}
\usepackage{booktabs}
\usepackage{enumitem}
\usepackage{bbding}
\usepackage{nicematrix}
\usepackage{nicefrac}       
\usepackage{microtype}      
\usepackage{footmisc}
\usepackage{wrapfig}
\usepackage{float}
\usepackage{multirow}
\usepackage{algorithm}
\usepackage[11pt]{moresize}
\usepackage{algpseudocode}
\usepackage{algorithmicx}  
\usepackage{algpseudocode}

\usepackage{caption}
\usepackage{amsfonts,amsthm,bbm,bm}
\usepackage{pifont}
\newcommand{\myhyperlink}[3][black]{\hyperlink{#2}{\color{#1}{#3}}}

\usepackage[pagebackref,breaklinks,colorlinks]{hyperref}

\usepackage[capitalize]{cleveref}
\newcommand{\dcircle}[1]{\ding{\numexpr181 + #1}}
\newcommand{\cmark}{\ding{51}}%
\newcommand{\xmark}{\ding{55}}%
\def\cvprPaperID{933} 
\def\confName{CVPR}
\def\confYear{2023}

\begin{document}

\title{Interactive Segmentation as Gaussian Process Classification}
\author{\hspace{-0.7cm}Minghao Zhou$^{1,2}$, Hong Wang$^{2,}$\footnotemark[1], Qian Zhao$^{1}$, Yuexiang Li$^{2}$, Yawen Huang$^{2}$, Deyu Meng$^{1,3,4}$, Yefeng Zheng$^{2}$\\
\hspace{-0.7cm}$^{1}$Xi'an Jiaotong University, Xi'an, China \quad
$^{2}$Tencent Jarvis Lab, Shenzhen, China \\
\hspace{-0.7cm}$^{3}$Peng Cheng Laboratory, Shenzhen, China \  
$^{4}$Macau University of Science and Technology, Macau, China\\
{\hspace{-0.7cm}\tt\small woshizhouminghao@stu.xjtu.edu.cn \  \{hazelhwang, vicyxli, yawenhuang, yefengzheng\}@tencent.com}  \\
{\hspace{-0.7cm}\tt\small\{timmy.zhaoqian, dymeng\}@mail.xjtu.edu.cn}
}
\maketitle
\renewcommand{\thefootnote}{\fnsymbol{footnote}}
\footnotetext[1]{Corresponding author}
\renewcommand{\thefootnote}{\arabic{footnote}}

\begin{abstract}

Click-based interactive segmentation (IS) aims to extract the target objects under user interaction. For this task, most of the current deep learning (DL)-based methods mainly follow the general pipelines of semantic segmentation. Albeit achieving promising performance, they do not fully and explicitly utilize and propagate the click information, inevitably leading to unsatisfactory segmentation results, even at clicked points. Against this issue, in this paper, we propose to formulate the IS task as a Gaussian process (GP)-based pixel-wise binary classification model on each image. To solve this model, we utilize amortized variational inference to approximate the intractable GP posterior in a data-driven manner and then decouple the approximated GP posterior into double space forms for efficient sampling with linear complexity. Then, we correspondingly construct a GP classification framework, named GPCIS, which is integrated with the deep kernel learning mechanism for more flexibility. The main specificities of the proposed GPCIS lie in: 1) Under the explicit guidance of the derived GP posterior, the information contained in clicks can be finely propagated to the entire image and then boost the segmentation; 2) The accuracy of predictions at clicks has good theoretical support. These merits of GPCIS as well as its good generality and high efficiency are substantiated by comprehensive experiments on several benchmarks, as compared with representative methods both quantitatively and qualitatively. Codes will be released at \href{https://github.com/zmhhmz/GPCIS_CVPR2023}{https://github.com/zmhhmz/GPCIS\_CVPR2023}.

\end{abstract}


\section{Introduction}
\label{sec:intro}

Driven by the huge potential in reducing the pixel-wise annotation cost, interactive segmentation (IS) has sparked much research interest~\cite{he2014interactive}, which aims to segment the target objects under user interaction with less interaction cost. Among various types of user interaction~\cite{xu2017deep,acuna2018efficient,castrejon2017annotating,ling2019fast,bai2014error,wu2014milcut,majumder2020two,zhang2020interactive}, in this paper, we focus on the popular click-based mode, where positive annotations are clicked on the target object while negative ones are clicked in the background regions~\cite{jang2019interactive,sofiiuk2020f,sofiiuk2021reviving,lin2022focuscut,chen2022focalclick}.


\begin{figure}[t]
  \centering
\includegraphics[width=1\linewidth]{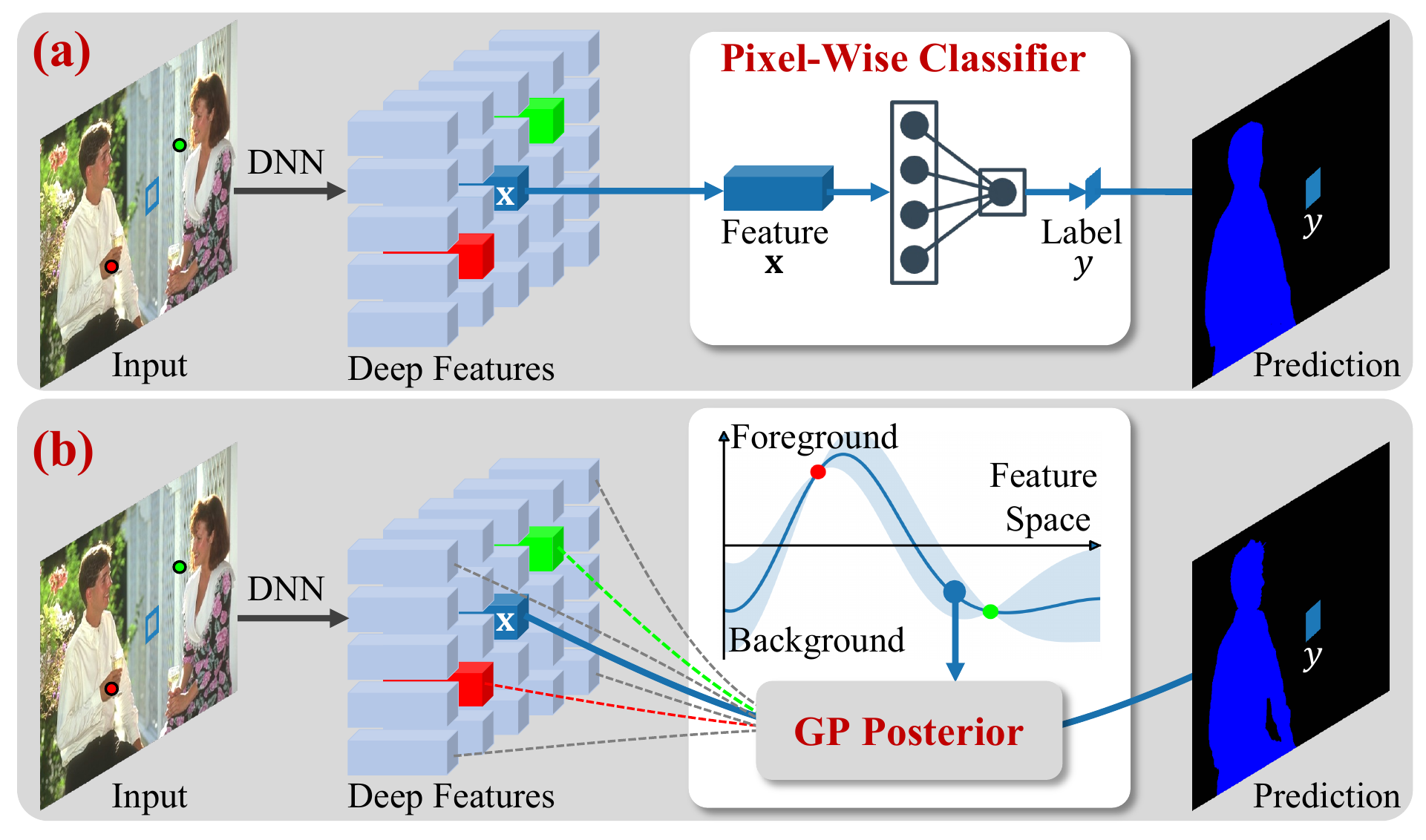}
  \vspace{-0.6cm}
  \caption{
 Classification procedure for an exemplar unclicked pixel (blue box) in the IS task. 
  (a) Most current deep learning-based IS methods individually perform pixel-wise classification on the deep feature $\textbf{x}$
  (b) We formulate the IS task as a Gaussian process (GP) classification model on each image, where red (green) clicks are viewed as training data with foreground (background) labels, and the unclicked pixel as the to-be-classified testing data. 
Based on the derived GP posterior inference framework, the relations between the deep feature $\textbf{x}$ of the testing pixel (blue solid line) and that of other pixels (dashed lines) can be finely modeled and then the information at clicks can be propagated to the entire image for improved prediction. }

  \label{fig:first}
  \vspace{-0.6cm}
\end{figure}

Recent years have witnessed the promising success of deep learning (DL)-based methods in the IS task. The most commonly adopted research line is that the user interaction is encoded as click maps and fed into a deep neural network (DNN) together with input images to extract deep features for the subsequent segmentation~\cite{xu2016deep,sofiiuk2021reviving}. However, these methods generally suffer from two limitations: 
1) As shown in Fig. \ref{fig:first} (a), after extracting the deep features, they generally perform pixel-wise classification without specific designs for the IS task. 
As a result, during the last-layer classification, 
the deep features of different pixels are not fully interactive and the information contained in clicked pixels cannot be finely propagated to other pixels under explicit regularization. 
2) There is no explicit theoretical support that the clicked regions can be properly activated and correctly classified.
Although some researchers have proposed different strategies, \eg, non-local-based modules~\cite{chen2021conditional} and the backpropagating refinement scheme~\cite{jang2019interactive,sofiiuk2020f}, they usually incur extra computational cost and are not capable enough to deal with the two problems simultaneously. Besides, the relations between deep features of different pixels are generally 
characterized and 
captured based on off-the-shelf network modules. Such implicit design makes it hard to clearly understand the working mechanism underlying these methods.

To alleviate these aforementioned issues, inspired by the intrinsic capabilities of {Gaussian} process (GP) models, \emph{e.g.}, explicitly measuring the relations between data points by a kernel function, and promoting accurate predictions at training data via interpolation, we rethink the IS task and attempt to construct a GP-based inference framework for the specific IS task. Concretely, as shown in Fig.~\ref{fig:first}~(b), we propose to treat the IS task from an alternative perspective and reformulate it as a pixel-level binary classification problem on each image, where clicks are viewed as training pixels with classification labels, \emph{i.e.}, foreground or background, and the unclicked points as the to-be-classified testing pixels. With such understanding, we construct the corresponding GP classification model. To solve it, we propose to utilize the amortized variational inference to efficiently approximate the intractable GP posterior in a data-driven manner, and then adopt the decoupling techniques~\cite{wilson2020efficiently,wilson2021pathwise} to achieve the GP posterior sampling with linear complexity. To improve the learning flexibility, we further embed the deep kernel learning strategy into the decoupled GP posterior inference procedure. Finally, by correspondingly integrating the derived GP posterior sampling mechanism with DNN backbones, we construct a GP Classification-based Interactive Segmentation framework, called GPCIS. In summary, our contributions are mainly three-fold:

\vspace{0.7mm}
\noindent 1) We propose to carefully formulate the IS task as a Gaussian process classification model on each image. To adapt the GP model to the IS task, we propose specific designs and accomplish the approximation and efficient sampling of the GP posterior, which are then effectively integrated with the deep kernel learning mechanism for more flexibility. 

\vspace{0.7mm}

\noindent 2) We build a concise and clear interactive segmentation network under a theoretically sound framework. 
As shown in Fig.~\ref{fig:first}~(b), the correlation between the deep features of different pixels is modeled by GP posterior. With such explicit regularization, the information contained in clicks can be finely propagated to the entire image and boost the prediction of unclicked pixels. Besides, our method can provide rational theoretical support for accurate predictions at clicked points. These merits are finely validated in Sec.~\ref{sec:verf}.

\vspace{0.7mm}
\noindent 3) Extensive experimental comparisons as well as model verification comprehensively substantiate the superiority of our proposed GPCIS in segmentation quality and interaction efficiency. It is worth mentioning that the proposed GPCIS can consistently achieve superior performance under different backbone segmentors, showing its fine generality.

\section{Related Work}

In this section, we briefly review the related work on the click-based interactive segmentation (IS) task. 

Traditional methods for IS \cite{grady2006random,gulshan2010geodesic,kim2010nonparametric,rother2004grabcut} generally utilize the low-level features of to-be-segmented images and build optimization-based graphical models, which usually suffer from unsatisfactory performance and low efficiency. Motivated by the promising success of deep neural networks (DNN) \cite{long2015fully,chen2017deeplab} in semantic segmentation, various methods have borrowed these pipelines for handling the IS task by transforming user interactions into click maps and taking them as the network input~\cite{xu2016deep,liew2017regional,li2018interactive,lin2020interactive}. In 99\%AccuracyNet \cite{forte2020getting} and RITM \cite{sofiiuk2021reviving}, the mask predicted during the previous click was also regarded as the network input for helping the predictions for the current click. Recently, to better exploit the information contained in clicks and further propagate it to the entire image for promoting the segmentation of unclicked points, FCANet \cite{lin2020interactive} put more emphasis on leveraging the first click and CDNet \cite{chen2021conditional} designed the non-local-based conditional diffusion modules. Although these methods can deal with the relations between the features of different pixels to some extent, they can hardly provide any explicit theoretical basis for corrected predictions at clicked points. To this end, BRS \cite{jang2019interactive}, f-BRS \cite{sofiiuk2020f}, and CA \cite{kontogianni2020continuous} have proposed to perform loss backpropagation during testing to adapt click maps or their network parameters to the current testing image. Clearly, the extra computation cost would adversely affect the efficiency of interactive segmentation. Recently, another research line, \eg, RIS-Net \cite{liew2017regional}, FocalClick \cite{chen2022focalclick}, and FocusCut \cite{lin2022focuscut}, deals with the IS task from a local view to refine the segmentation results. Albeit attaining performance improvement, these methods have not fully exploited the relations between the deep features of clicks and those of unclicked points. 
Against this issue, in this paper, we build a concise and efficient model to explicitly model the relations between the deep features of the entire to-be-segmented image. 
It is worth noting that \cite{triebel2014active} employs a Gaussian process model to develop an active learning framework for interactive segmentation, aiming to actively query pixels to be labeled.


\section{Preliminaries: Gaussian Processes}\label{sec:pre}
\label{sec:gp}
Gaussian processes (GP) \cite{williams2006gaussian} can be understood as the ``Gaussian distribution over functions". As a compelling tool, by directly modeling the prior and posterior of functions, it has been widely adopted in various tasks \cite{williams1995gaussian,nickisch2008approximations,liu2020gaussian}. Mathematically, a GP is defined as a stochastic process where the joint distribution of any finite random variables is Gaussian. Define a mean function $\mu:\mathcal{X}\to \mathbb{R}$ and a covariance function $k: \mathcal{X}\times \mathcal{X}\to \mathbb{R}$, a GP $f\sim \mathcal{GP}(\mu,k)$ satisfies $\mathbf{f}_n = [f(\mathbf{x}_1),\cdots,f(\mathbf{x}_n)]^T \sim \mathcal{N}(\bm{\mu}_n,\mathbf{K}_{n,n})$ with mean $\bm{\mu}_n=[\mu(\mathbf{x}_1),\cdots,\mu(\mathbf{x}_n)]^T$ and covariance matrix $\mathbf{K}_{n,n}=k(\mathbf{X}_n,\mathbf{X}_n)\triangleq \{k(\mathbf{x}_i,\mathbf{x}_j)\}_{ij}$, for any finite observations $\mathbf{X}_n=[\mathbf{x}_1,\cdots,\mathbf{x}_n]^T \in \mathcal{X}^n$. Specifically, for the GP prior, $\mu(\cdot)$ is generally assumed to be a constant zero function. The covariance function $k(\cdot, \cdot)$ can be elaborately designed to model the correlations between the data points.



Given $n$ noise-free latent observations $ \mathbf{f}_n$ at training data $\mathbf{X}_n$, the GP posterior at testing data $\mathbf{X}_*$ is written as \cite{williams2006gaussian}: \vspace{-2mm}
\begin{equation} \label{eqn:gp1}
\mathbf{f}_*|\mathbf{X}_*,\mathbf{X}_n, \mathbf{f}_n \sim \mathcal{N}(\bm{\mu}_{*|n}, \mathbf{K}_{*,*|n}), \vspace{-2mm}
\end{equation}
where \vspace{-2mm}
\begin{equation} \label{eqn:gp2}
\bm{\mu}_{*|n}\!=\!\mathbf{K}_{*,n} \mathbf{K}_{n,n}^{-1} \mathbf{f}_n,\ \ \mathbf{K}_{*,*|n}\!=\! \mathbf{K}_{*,*} \!-\! \mathbf{K}_{*,n} \mathbf{K}_{n,n}^{-1}\mathbf{K}_{n,*}.\! 
\end{equation}

As seen, the GP posterior utilizes the relations between the testing data $\mathbf{X}_*$ and the training data $\mathbf{X}_n$ to estimate the distribution of the function $f$ at $\mathbf{X}_*$, where the relations 
are explicitly measured by the kernel function $k(\cdot,\cdot)$.



\section{Methodology}\label{sec:method}

In this section, we firstly propose that the interactive segmentation (IS) problem can be regarded as a pixel-wise binary classification task on each input image. Based on such understanding, we carefully formulate this task with a GP classification model. Then, to solve it, we propose the corresponding algorithms to finely approximate and efficiently sample from GP posterior. Finally, by flexibly combining the GP model with DNN backbones, we construct the entire inference framework. The details are given below.

 \subsection{Model Formulation}
For the interactive segmentation on an RGB image $\mathcal{I}\in \mathbb{R}^{m\times 3}$, users iteratively impose positive or negative clicks $\{c,y_c\}_{c=1}^n$ on the image to segment the target object, where $m$ is the number of the pixels; $n$ is the number of the interactive clicks; and $y_c\in\{1,-1\}$ is the label (\emph{i.e.}, foreground/background) of the $c^{\text{th}}$ click. 
By feeding the to-be-segmented image $\mathcal{I}$ to a DNN $g_\psi(\cdot)$, we can extract the feature representations as $g_\psi(\mathcal{I})=\mathbf{X}_{m}=[\mathbf{x}_1,\cdots,\mathbf{x}_m]^T\in \mathbb{R}^{m\times d}$, where $\mathbf{x}_i\in \mathbb{R}^d$ denotes the features of pixel $i$. Given the features at clicked pixels $\mathbf{X}_n=[\mathbf{x}_1,\cdots,\mathbf{x}_n]^T\in\mathbb{R}^{n\times d}$ and their labels $\mathbf{y}_n\in\{1,-1\}^n$, our goal is to predict the labels $\mathbf{y}_*$ of the unclicked pixels with the features $\mathbf{X}_*\in\mathbb{R}^{*\times d}$, where $*=m-n$ is the number of unclicked pixels. Next, we aim to solve the two core problems: \hypertarget{R1}{\dcircle{1}} How to finely model the relations between the deep features of different pixels and fully propagate the information contained in clicks for boosting the correct predictions at unclicked pixels? \hypertarget{R2}{\dcircle{2}} How to guide and promote accurate predictions at clicks?

Inspired by the appealing properties of Gaussian process (GP) models for our task, \eg, the capability of explicitly modeling the relations between data points and accurately interpolating the training data, we propose to rethink the IS task from a micro perspective and formulate it as a pixel-level binary classification task on each image, where the features of clicked pixels $\mathbf{X}_n$ are regarded as training data with labels $\mathbf{y}_n$ and those of unclicked pixels $\mathbf{X}_*$ as testing data. Based on such understanding, we attempt to handle the pixel-wise binary classification task via GP models. 

Specifically, we define a GP with a zero-mean prior $\mu(\cdot)$ and a covariance function $k(\cdot,\cdot)$ over the classification function $f:\mathbb{R}^d \to \mathbb{R}$, which takes the feature $\mathbf{x}_{i}$ of pixel $i$ as input and outputs the score for binary classification, \emph{i.e.}, positive score for foreground and negative score for background. Then the inference process from the available click information $\{\mathbf{X}_n,\mathbf{y}_n\}$ to the unknown labels $\mathbf{y}_*$ at $\mathbf{X}_*$ can be transformed into the following GP classification model which aims to solve the posterior distribution of the labels $\mathbf{y}_*$ given $\{\mathbf{X}_*, \mathbf{X}_n,\mathbf{y}_n\}$, mathematically expressed as: 
\vspace{0mm} 
\begin{equation} \label{eqn:result} 
p(\mathbf{y}_*|\mathbf{X}_*,\mathbf{X}_n,\mathbf{y}_n)\! =\! \int\! p(\mathbf{y}_*|\mathbf{f}_*)p(\mathbf{f}_*|\mathbf{X}_*,\mathbf{X}_n,\mathbf{y}_n) \mathrm{d}\mathbf{f}_*, 
\end{equation} 
where $p(\mathbf{f}_*|\mathbf{X}_*,\!\mathbf{X}_n,\!\mathbf{y}_n)$ is the \textbf{GP posterior}. For the binary classification task, it is conventionally set that $p(\mathbf{y}_*|\mathbf{f}_*)=\Pi_{u=1}^* s(y_u f_u)$, where $s(\cdot)$ is the sigmoid function~\cite{nickisch2008approximations}. 

As seen, the integral in Eq.~\eqref{eqn:result} is explicitly intractable for our task. Fortunately, if we can obtain the GP posterior, the integral can be approximated with a Monte Carlo method \cite{hensman2015scalable}. Specifically, suppose $\tilde{\mathbf{f}}_*$ is sampled from the derived GP posterior, we can approximately get that the probability of classifying the testing data $\mathbf{X}_*$ into the foreground is $\tilde{\mathbf{y}}_*=s(\tilde{\mathbf{f}}_*)$. Hence, the key is how to obtain the GP posterior $p(\mathbf{f}_*|\mathbf{X}_*,\!\mathbf{X}_n,\!\mathbf{y}_n)$. Besides, after obtaining the GP posterior, how to achieve efficient sampling from it is also worth exploring since high inference efficiency is critical  for the IS task. Next, we will answer the two questions.

\vspace{1mm}
\subsection{GP Posterior Approximation and Sampling}
In this subsection, we aim to  approximate the GP posterior and achieve efficient sampling.  
 
\vspace{1mm}
\noindent\textbf{GP posterior approximation.} 
It is easily known that the GP posterior $p(\mathbf{f}_*|\mathbf{X}_*,\!\mathbf{X}_n,\!\mathbf{y}_n)$ can be rewritten as: 
\begin{equation} \label{eqn:int}
\hspace{-1mm}p(\mathbf{f}_*|\mathbf{X}_*,\!\mathbf{X}_n,\!\mathbf{y}_n) \!=\!\!\! \int \!\!p(\mathbf{f}_*|\mathbf{X}_*,\!\mathbf{X}_n,\!\mathbf{f}_n) p(\mathbf{f}_n|\mathbf{X}_n,\mathbf{y}_n)\mathrm{d}\mathbf{f}_n,\! 
\end{equation}
where $p(\mathbf{f}_*|\mathbf{X}_*,\!\mathbf{X}_n,\!\mathbf{f}_n)$ follows a Gaussian distribution as defined in Eq. (\ref{eqn:gp1}); $p(\mathbf{f}_n|\mathbf{X}_n,\mathbf{y}_n)\propto p(\mathbf{y}_n|\mathbf{X}_n,\mathbf{f}_n)p(\mathbf{f}_n|\mathbf{X}_n)$; and $p(\mathbf{f}_n|\mathbf{X}_n) = \mathcal{N}(\bm{\mu}_n,\mathbf{K}_{n,n})$.
For the classification task, due to the non-Gaussian likelihood $p(\mathbf{y}_n|\mathbf{X}_n,\mathbf{f}_n)=\Pi_{c=1}^n s(y_c f_c)$, $p(\mathbf{f}_n|\mathbf{X}_n,\mathbf{y}_n)$ is non-Gaussian and leads to that the GP posterior $p(\mathbf{f}_*|\mathbf{X}_*,\!\mathbf{X}_n,\!\mathbf{y}_n)$ in Eq. (\ref{eqn:int}) is intractable.
Against this issue, previous methods~\cite{opper2009variational,nickisch2008approximations,hensman2015scalable} have proposed to approximate $p(\mathbf{f}_n|\mathbf{X}_n,\mathbf{y}_n)$ with a Gaussian variational distribution $q(\mathbf{f}_n|\mathbf{X}_n,\mathbf{y}_n)$ by minimizing their KL divergence as: 
\begin{equation}\label{eqn:kl}
    \min_q D_{\mathit{KL}}(q(\mathbf{f}_n|\mathbf{X}_n,\mathbf{y}_n)||p(\mathbf{f}_n|\mathbf{X}_n,\mathbf{y}_n)). 
\end{equation}
To solve Eq. (\ref{eqn:kl}),  conventional variational inference-based methods \cite{opper2009variational,nickisch2008approximations,hensman2015scalable} independently optimize the objective on each training task (\emph{i.e.}, each training image in our IS case). These methods are generally time-consuming and fail to exploit the shared information among different images. In contrast, we imitate the amortized variation inference \cite{kingma2013auto} to efficiently infer $q(\mathbf{f}_n|\mathbf{X}_n,\mathbf{y}_n)$ from $\{\mathbf{X}_n, \mathbf{y}_n\}$ and the distribution parameters for $q(\mathbf{f}_n|\mathbf{X}_n,\mathbf{y}_n)$ can be flexibly learned based on all the training images (\emph{i.e.}, the whole benchmark dataset) in an end-to-end manner. Specifically, the variational distribution $q(\mathbf{f}_n|\mathbf{X}_n,\mathbf{y}_n)$ is set as:
\begin{equation}\label{my1}
q(\mathbf{f}_n|\mathbf{X}_n,\mathbf{y}_n)=\mathcal{N}(\mathbf{m}_\xi(\mathbf{X}_n,\mathbf{y}_n),\sigma^2 \mathbf{I}_n),  
\end{equation}
where the mean function $\mathbf{m}_\xi(\mathbf{X}_n,\mathbf{y}_n)$ is designed as: 
\begin{equation}\label{my2}
\mathbf{m}_\xi(\mathbf{X}_n, \mathbf{y}_n) = \mathrm{Softplus}(\mathrm{MLP}_\xi(\mathbf{X}_n)) * \mathbf{y}_n, 
\end{equation}
where MLP$_\xi(\cdot)$ represents a multi-layer perceptron parameterized by $\xi$, which transforms the features $\mathbf{X}_n$ from $\mathbb{R}^{n\times d}$ to $\mathbb{R}^{n\times 1}$. The activation function $\mathrm{Softplus(x)=\log (1+e^x)}$ is the smooth version of ReLU, whose output is consistently positive. 
By empirically setting a small variance $\sigma^2$ as 0.01, for any $\mathbf{f}_n\sim q(\mathbf{f}_n|\mathbf{X}_n,\mathbf{y}_n)$, we have $\mathbf{f}_n\approx \mathbf{m}_\xi$, which has the same positive/negative sign as $\mathbf{y}_n$ and helps the correct category prediction at clicks.

By substituting Eq.~\eqref{my1} and $p(\mathbf{f}_n|\mathbf{X}_n,\mathbf{y}_n)$ derived in Eq.~\eqref{eqn:int}, we can rewrite the KL divergence in Eq.~\eqref{eqn:kl} as:~\footnote{Please refer to \textit{Supplementary Material} (\textit{SM}) for detailed derivations. \label{ft2}}
\begin{align} 
\min_{\xi}&-\!\mathbb{E}_{q(\mathbf{f}_n|\mathbf{X}_n,\mathbf{y}_n)\sim \mathcal{N}(\mathbf{m}_{\xi},\sigma^2 \mathbf{I}_n)}\textstyle{\sum_{c=1}^n}\big[  y_c\log s(f_c) \nonumber \\ 
&\! +\! (1 - y_c)\log(1 - s(f_c)) \big] + \textstyle{\frac{1}{2}}\mathbf{m}_\xi^T\mathbf{K}_{n,n}^{-1}\mathbf{m}_\xi,  \label{eqn:kl2} 
\end{align} 
where we simplify $\mathbf{m}_{\xi}(\mathbf{X}_n,\mathbf{y}_n)$ as $\mathbf{m}_\xi$.

By optimizing Eq.~\eqref{eqn:kl2} over all the training images in an end-to-end manner, we can obtain the variational distribution $q(\mathbf{f}_n|\mathbf{X}_n,\mathbf{y}_n)\!=\!\mathcal{N}(\mathbf{m}_\xi,\sigma^2 \mathbf{I}_n)$. Then by substituting it into Eq. (\ref{eqn:int}), we can easily derive that the GP posterior $p(\mathbf{f}_*|\mathbf{X}_*,\!\mathbf{X}_n,\!\mathbf{y}_n)$ is Gaussian and can be approximated as: \footref{ft2} 
\begin{equation}\label{my3}
p(\mathbf{f}_*|\mathbf{X}_*,\!\mathbf{X}_n,\!\mathbf{y}_n) \sim \mathcal{N}(\bm{\mu}_{*|n}, \mathbf{K}_{*,*|n}), \vspace{-2mm}
\end{equation}
where \vspace{-2mm}
\begin{equation} \label{eqn:gp2}
\begin{split}
\bm{\mu}_{*|n}&=\mathbf{K}_{*,n} \mathbf{K}_{n,n}^{-1}  \mathbf{m}_\xi, \\
\mathbf{K}_{*,*|n} &= \mathbf{K}_{*,*} - \mathbf{K}_{*,n}\mathbf{K}_{n,n}^{-1} (\mathbf{I}_{n}-{\sigma^2}\mathbf{K}_{n,n}^{-1})\mathbf{K}_{n,*}.  \vspace{-2mm}
\end{split}
\end{equation}

\begin{figure*}[t]
  \centering
  \vspace{-0.7cm}
\includegraphics[width=0.84\linewidth]{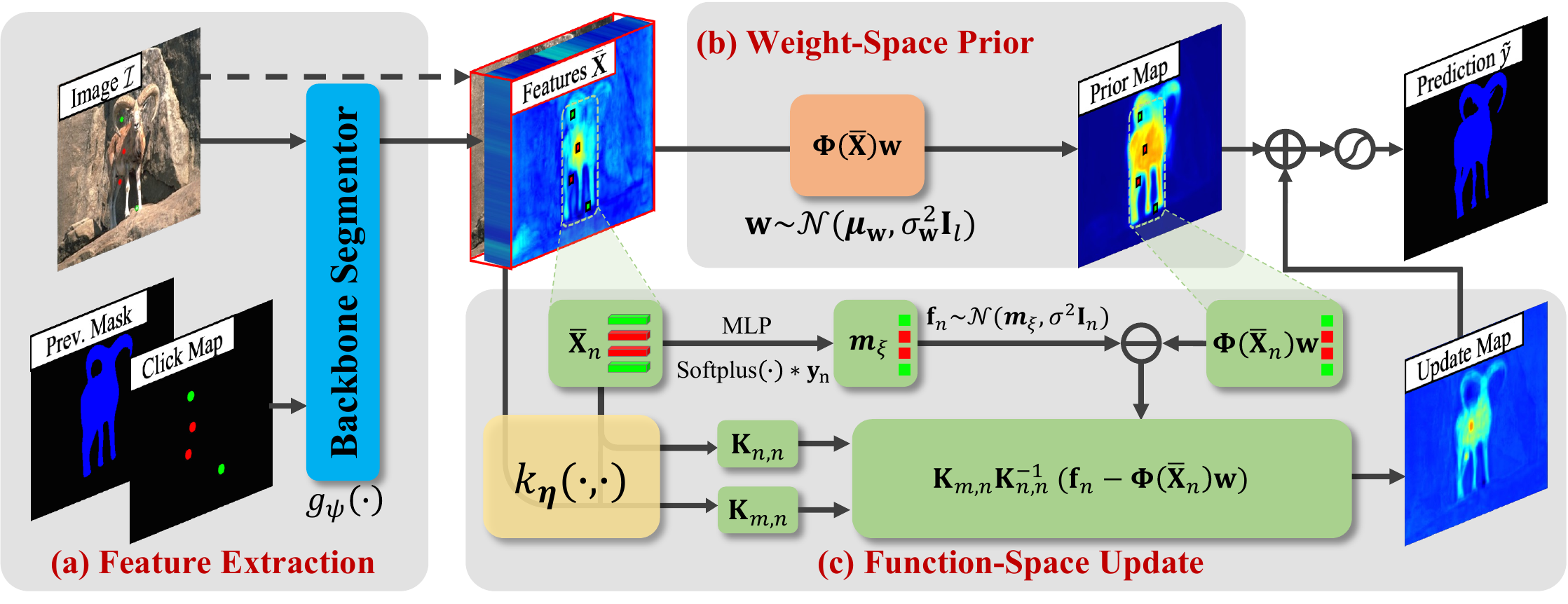}
  \vspace{-0.25cm}
  \caption{{\textcolor{black}{The general framework of the proposed Gaussian Process Classification-based Interactive Segmentation (GPCIS)}}. It consists of (a) an off-the-shelf backbone segmentor $g_{\psi}\left(\cdot\right)$ for extracting the deep features and (b)+(c) the GP posterior inference for predicting the segmentation result $\Tilde{y}$. Specifically, the GP posterior inference is composed of (b) the weight-space prior term and (c) the function-space update term, as derived in Eq.~\eqref{eqn:inference}. As seen, the proposed GPCIS is built under a theoretically sound framework.}
  \label{fig:main}
  \vspace{-0.3cm}
\end{figure*}



\noindent\textbf{Decoupling GP posterior for efficient sampling.}
From the analysis of Eq.~\eqref{eqn:result}, by sampling $\tilde{\mathbf{f}}_*$ from the tractable GP posterior $p(\mathbf{f}_*|\mathbf{X}_*,\!\mathbf{X}_n,\!\mathbf{y}_n)$ in Eq.~\eqref{my3}, we can obtain the classification probability for unclicked pixels as $\tilde{\mathbf{y}}_*=s(\tilde{\mathbf{f}}_*)$. To sample $\tilde{\mathbf{f}}_*$, the standard approach is to compute $\tilde{\mathbf{f}}_* = \mathbf{\mu}_{*|n} + \mathbf{K}_{*,*|n}^{1/2}\bm{\zeta}$ with $\bm{\zeta} \sim \mathcal{N}(0,\mathbf{I}_n)$~\cite{wilson2020efficiently}. As seen, the computation cost of $\mathbf{K}_{*,*|n}^{1/2}$ is cubic w.r.t. the number of unclicked pixels $*$, \emph{i.e.}, $\mathcal{O}(*^3)$, which severely affects the efficiency. 
Against this issue, we propose to adopt the techniques~\cite{wilson2020efficiently,wilson2021pathwise} which decouple the GP posterior into a weight-space prior term and a function-space update term, largely reducing the sampling cost without sacrificing interpolation accuracy at clicks. Then, for the GP posterior in Eq.~\eqref{my3}, we can derive the sampling framework as~\cite{wilson2020efficiently,wilson2021pathwise}:~\footref{ft2} 
\begin{equation} \label{eqn:inference}
\hspace{-2.5mm}\tilde{\mathbf{f}}_*  = \! \!\!\!\underbrace{\mathbf{\Phi}(\mathbf{X}_*) \mathbf{w}}_{\text{weight-space prior}}
\!\!\!\!+\underbrace{\mathbf{K}_{*,n}\mathbf{K}_{n,n}^{ -1}(\mathbf{f}_n-\mathbf{\Phi}(\mathbf{X}_n)\mathbf{w})}_{\text{function-space update}},\!\! 
\end{equation}
where $\mathbf{w}\sim \mathcal{N}(0,\mathbf{I}_l)$; {\textcolor{black}{$\mathbf{f}_n\sim q(\mathbf{f}_n| \mathbf{X}_n,\mathbf{y}_n)=\mathcal{N}(\mathbf{m}_{\xi},\sigma^2 \mathbf{I}_n)$}};  $\mathbf{\Phi}(\mathbf{X})=\{\phi_r(\mathbf{x}_i)\}_{ir}\in \mathbb{R}^{m\times l}$ is constructed by a set of $l$ Fourier bases and the $r$-th basis is expressed as~\cite{rahimi2007random}: 
\begin{equation} \label{eqn:rff}
\phi_r(\mathbf{x}) = \sqrt{2/l}\ \mathrm{cos}(\bm{\theta}_r^T \mathbf{x}+\tau_r), 
\end{equation}
where $i=1,2,\ldots,m$; $r=1,2,\ldots,l$;  $\tau_r\! \sim\! U(0,2\pi)$; $\bm{\theta}_r \in \mathbb{R}^d$ is sampled from the spectral density of the kernel $k(\cdot,\cdot)$. We will carefully design the kernel function in Sec.~\ref{sec:dkl}.

{\textcolor{black}{In practice, considering $l\ll *$ and $n\ll *$, the cost of sampling from Eq.~\eqref{eqn:inference} is reduced from $\mathcal{O}(*^3)$ to $\mathcal{O}(*)$~\cite{wilson2020efficiently,wilson2021pathwise}. Note that in our practical implementation, to keep consistency with most DL-based methods~\cite{sofiiuk2021reviving,chen2021conditional,chen2022focalclick}, we execute an inference on the entire image with $m$ pixels. That is to say, we also sample $\tilde{\mathbf{f}}_n$ using Eq. (\ref{eqn:inference}) in parallel with $\tilde{\mathbf{f}}_*$, by replacing the subscripts $*$ (\emph{i.e.}, the number of unclicked pixels) with the total number of pixels $m$. Then, we can obtain the entire prediction results of $m$ pixels, \ie, $\tilde{\mathbf{y}}=s(\tilde{\mathbf{f}}_m)$. }}

\vspace{2mm}


\noindent\hypertarget{remark1}{\textit{\textbf{Remark 1:}}} 
It is worth mentioning that the proposed sampling strategy in Eq.~\eqref{eqn:inference} possesses two inherent characteristics: \hypertarget{RR1}{\myhyperlink{R1}{\dcircle{1}}} The relations between the deep features of clicked points and those of the unclicked points are fully utilized and explicitly modeled by the function-space update term, which enables the information contained in clicked regions to propagate to other regions. 
\hypertarget{RR2}{\myhyperlink{R2}{\dcircle{2}}} 
For training stability, the matrix inversion $\mathbf{K}_{n,n}^{-1}$ in Eqs. (\ref{eqn:kl2}) (\ref{eqn:inference}) is practically computed by $(\mathbf{K}_{n,n}+\epsilon^2\mathbf{I})^{-1}$, where $\epsilon^2$ is empirically set to 0.01 during training. 
In Eq. (\ref{eqn:inference}), if we replace the number of the unclicked pixels (subscripts $*$) with the number of clicked pixels (subscripts $n$) and set a small enough $\epsilon^{2}$, we can obtain that $\tilde{\mathbf{f}}_n \approx \mathbf{f}_n\approx \mathbf{m}_\xi=\mathrm{Softplus}(\mathrm{MLP}_\xi(\mathbf{X}_n)) * \mathbf{y}_n$, showing that the sampled $\tilde{\mathbf{f}}_n$ has the same positive/negative sign as the labels $\mathbf{y}_n$. This implies that the proposed sampling strategy can provide theoretical support for encouraging accurate predictions at clicked points. 
The two characteristics are validated by the model verification in Sec.~\ref{sec:verf}.

\subsection{Double Space Deep Kernel Learning}\label{sec:dkl}
From Eqs.~\eqref{eqn:inference} and \eqref{eqn:rff}, we can see that the kernel $k(\cdot,\cdot)$ affects both the function-space update and weight-space prior terms. Designing a proper and flexible kernel is important for better modeling the relations between pixels and extracting the {\textcolor{black}{prior knowledge underlying the segmentation function}}.

In the decoupling paradigm~\cite{wilson2020efficiently,wilson2021pathwise}, the adopted kernel function is generally pre-defined and fixed, which would lead to two potential limitations: 
1) In function space, the kernel representation capacity would be constrained and the similarity measure between data points may not be optimal for our task; 
2) In weight space, the prior term is not flexible enough to capture the prior knowledge underlying the IS task.  
Against these issues, instead of adopting the fixed manually-designed kernels, inspired by deep kernel learning (DKL) \cite{wilson2016deep}, we propose to flexibly learn the kernel function in both function space and weight space from the abundant training images in a data-driven manner.

Specifically, we propose to perform double space DKL on $\Bar{\mathbf{x}}_i\in\mathbb{R}^{d+3}$, where $\Bar{\mathbf{x}}_i$ represents the concatenation of the deep features $\mathbf{x}_i \in\mathbb{R}^{d}$ and the image RGB values $\mathcal{I}_i \in\mathbb{R}^{3}$ at pixel $i$. Here, the concatenation of input image $\mathcal{I}$ is for providing more information as validated in Sec.~\ref{sec:abla}. In function space, to improve representation flexibility, we select a modified radial basis function (RBF) with scaling hyperparamters $\bm{\eta} = \{\eta_0,\cdots,\eta_d\}$ as the kernel function: 
{\textcolor{black}{\(
k_{\bm{\eta}}(\Bar{\mathbf{x}}_i,\Bar{\mathbf{x}}_j) = \eta_0\ \mathrm{exp}(-\sum_{t=1}^3 (\mathcal{I}_{it}-\mathcal{I}_{jt})^2/2) + \mathrm{exp}(-\sum_{t=1}^d (\mathbf{x}_{it}-\mathbf{x}_{jt})^2/(2\eta_t)),\)}}
where $\forall t, \eta_t>0$ and $\mathbf{x}_{it}$ is the $t$-th element of $\mathbf{x}_{i}$. In weight space, since the hyperparameters $\bm{\eta}$ are updating during network training, it is not suitable to sample $\bm{\theta}_r$ in Eq.~\eqref{eqn:rff} from the kernel's spectral density, thus it is set as a learnable parameter. To further improve flexibility and representation capacity of the weight-space prior term for better extracting the image prior, we parameterize the prior distribution of $\mathbf{w}$ as $\mathbf{w}\sim\mathcal{N}(\bm{\mu}_{\mathbf{w}}, \sigma^2_{\mathbf{w}}\mathbf{I}_l)$. These hyperparameters in the double space, including ${\bm{\eta}}, \bm{\theta},\bm{\tau}, \bm{\mu}_{\mathbf{w}}$, and $\sigma^2_{\mathbf{w}}$, are trained in an end-to-end manner based on the entire training dataset. 

Compared to the pre-fixed kernel design manner, the proposed double space DKL strategy is more flexible and it can utilize the powerful representation ability of DNNs to promote the performance, which is validated in Sec.~\ref{sec:abla}.

\begin{figure*}[t]
  \centering
  \vspace{-2mm}
\includegraphics[width=0.98\linewidth]{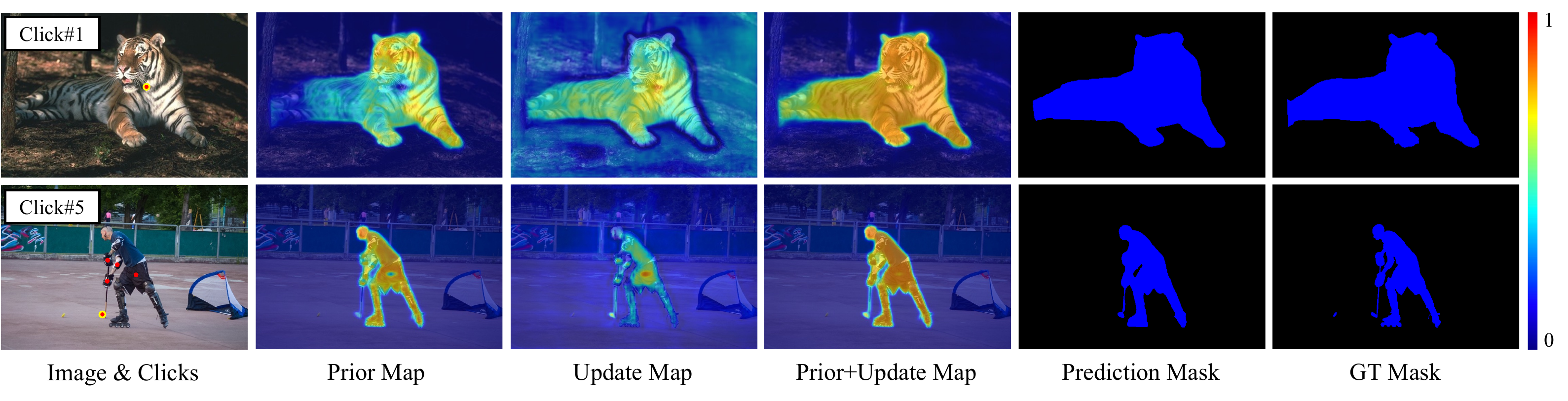}
  \vspace{-0.4cm}
\caption{Visual verification of GPCIS's working mechanism, including probability maps of weight-space prior and function-space update.}
  \label{fig:vis}
  \vspace{-0.4cm}
\end{figure*}

\subsection{The Proposed GPCIS Framework}
Based on the derived GP posterior sampling procedure as well as the double space DKL mechanism, 
we can correspondingly construct the entire framework, called Gaussian Process Classification-based Interactive Segmentation (GPCIS). As presented in Fig.~\ref{fig:main}, similar to~\cite{sofiiuk2021reviving,chen2021conditional,chen2022focalclick}, we firstly input the image $\mathcal{I}$ and the click maps together with the previous mask to a general backbone segmentor $g_{\psi}\left(\cdot\right)$ for extracting the deep features $\mathbf{X}$. By feeding the concatenation of $\mathbf{X}$ and the image $\mathcal{I}$, \ie, $\Bar{\mathbf{X}}$, to the efficient GP posterior sampling framework in Eq.~(\ref{eqn:inference}), we can generate a weight-space prior map and a function-space update map. Finally, we can obtain the segmentation result $\tilde{\mathbf{y}}$ by adding the two maps followed by a sigmoid function.
 
\vspace{1mm}
\noindent \hypertarget{remark2}{\textbf{\textit{Remark 2:}}}
As seen, in our proposed GPCIS, the correlation modeling on the deep features of different pixels are explicitly corresponding to the derived GP posterior sampling strategy. Compared to the current methods \cite{jang2019interactive,sofiiuk2020f,chen2021conditional,lin2022focuscut,chen2022focalclick} which are implicitly built based on off-the-shelf network modules, our method has a clearer working mechanism.

\vspace{1mm}
\noindent \textbf{Network training.} 
For the proposed GPCIS framework, the involved parameters are automatically learned from the training data in an end-to-end manner, including $\psi$ for the backbone segmentor, $\xi$ for variational distribution $q(\mathbf{f}_n|\mathbf{X}_n,\mathbf{y}_n)$, $\bm{\eta}$ for function-space DKL, and $\{\bm{\theta}, \bm{\tau}, \bm{\mu}_{\mathbf{w}},\sigma_{\mathbf{w}}\}$ for weight-space DKL. 
The training loss $\mathcal{L}$ is set as:~\footnote{The entire algorithm flowchart is provided in \textit{SM}.} 
\begin{equation}
\mathcal{L} = \mathcal{L}_{\mathit{NFL}}(\tilde{\mathbf{y}},\mathbf{y}_{\mathit{gt}}) + \alpha \ \mathcal{L}_{\mathit{VI}},
\end{equation}
where $\tilde{\mathbf{y}}$ is the output segmentation result; $\mathbf{y}_{\mathit{gt}}$ is the ground truth mask; $\alpha$ is the weighting parameter which is empirically set to $10^{-3}$;  $\mathcal{L}_{\mathit{NFL}}$ is the normalized focal loss~\cite{sofiiuk2019adaptis} which is widely adopted by the existing IS methods~\cite{sofiiuk2021reviving,chen2021conditional,chen2022focalclick};  $\mathcal{L}_{\mathit{VI}}$ is the optimization objective in Eq. (\ref{eqn:kl2}).

\section{Experiments}

\subsection{Experimental Settings}

\noindent \textbf{Datasets.} We conduct IS experiments on four widely-adopted datasets: 1) \textbf{GrabCut} \cite{rother2004grabcut} contains 50 images with single object masks; 2) \textbf{Berkeley} \cite{mcguinness2010comparative} contains 96 images with 100 object masks; 3) \textbf{SBD} \cite{hariharan2011semantic} contains 20,172 masks for 8,498 images as the training set, 6,671 instance-level masks for 2,857 images as the validation set. The annotated masks are polygonal; 4) \textbf{DAVIS} \cite{perazzi2016benchmark} contains 345 frames randomly sampled from 50 videos, with high-quality masks. We adopt the training split of SBD as the training set and take the other mentioned datasets for evaluation.

\vspace{1mm}
\noindent \textbf{Evaluation metrics.}
Following~\cite{xu2016deep,li2018interactive,sofiiuk2021reviving,lin2022focuscut,chen2022focalclick}, we adopt the same strategy to simulate the clicks, which generates the next click at the center of the largest error region by comparing the prediction and ground truth. The Number of Clicks (NoC) is adopted as the metric, which counts the average number of clicks needed to achieve the target Intersection over Union (IoU).  Following~\cite{xu2016deep,li2018interactive,sofiiuk2021reviving,lin2022focuscut,chen2022focalclick}, we set the IoU threshold to 85\% and 90\%. The evaluation metrics are denoted as NoC@85 and NoC@90, respectively. The default maximum number of clicks $n$ is 20. The Number of Failures (NoF) is also reported and it counts the number of images that cannot achieve the target IoU within the specified maximum number of clicks. Besides,
we also report the average IoU at the N-th click, denoting IoU\&N. To evaluate the correctness of predictions at clicks, we propose a new metric as NoIC which counts the Number of Incorrectly classified Clicks over a testing dataset. Lower NoC, NoF, and NoIC, as well as higher IoU\&N, indicate better performance.

\vspace{1mm}
\noindent \textbf{Implementation details.} 
We implement the proposed framework with PyTorch~\cite{paszke2019pytorch} based on 4 NVIDIA V100 GPUs. For the backbone segmentor, we adopt three different networks, including SegFormerB0-S2 \cite{xie2021segformer,chen2022focalclick}, HRNet18s-S2 \cite{wang2020deep,chen2022focalclick}, and DeepLabv3+~\cite{chen2018encoder} with ResNet50~\cite{he2016deep}, to substantiate the generality of our method. The initial learning rate is $5\times 10^{-3}$ for SegFormerB0-S2 and ResNet50, and $5\times 10^{-4}$ for HRNet18s-S2. It is divided by 10 at [190, 220] epochs and the total number of epochs is 230, as in \cite{chen2022focalclick}. We adopt the Adam optimizer \cite{kingma2014adam} with the total batch size of 64 and the training patch size of $256\times 256$. For inferring $\mathbf{m}_\xi$ in Eq.~\eqref{my1}, we adopt a one-hidden-layer MLP with 96 hidden units. More details are provided in \textit{SM}. 

\begin{table}[t] 
\footnotesize
\caption{The effect of $\epsilon^2$ on the NoIC of our proposed GPCIS with the backbone segmentor ResNet50 on the DAVIS dataset\cite{perazzi2016benchmark}. \vspace{-3mm}}
\label{tab:sigma}
\centering
\setlength{\tabcolsep}{1.8mm}{
\begin{tabular}{@{}c|c|c|c|c|c|c|c@{}}
\toprule 
  $\epsilon^2$ & $10^{-1}$ & $10^{-2}$ & $10^{-3}$ & $10^{-4}$ & $10^{-5}$ & $10^{-6}$ & $10^{-7}$  \\ \hline
  NoIC & 36 & 30 & 21 & 15  &15 & 8 & 2  \\
 \bottomrule
\end{tabular}}
\vspace{-5mm}
\end{table}

\begin{table*}[t] 
\footnotesize
\vspace{-0.2cm}
\caption{NoC@85 and NoC@90 of different competing methods on four datasets, \emph{i.e.}, GrabCut, Berkeley, SBD, and DAVIS. `*' denotes the models trained on the Augmented PASCAL VOC dataset \cite{everingham2010pascal,hariharan2011semantic}. Bold and underlined results indicate the top $1^{\text{st}}$ and $2^{\text{nd}}$ rank, respectively.}\vspace{-3mm}

\label{tab:exp}
\centering
\renewcommand{\arraystretch}{0.94}
\setlength{\tabcolsep}{0.7mm}{
\begin{tabular}{@{}c|l|cc|cc|cc|cc|cc@{}}
\toprule \multicolumn{1}{c|}{\multirow{2}{*}{Backbone}} &
\multicolumn{1}{c|}{\multirow{2}{*}{Method}} & \multicolumn{2}{c|}{GrabCut \cite{rother2004grabcut}} & \multicolumn{2}{c|}{Berkeley \cite{mcguinness2010comparative}} & \multicolumn{2}{c|}{SBD \cite{hariharan2011semantic}} & \multicolumn{2}{c|}{DAVIS \cite{perazzi2016benchmark} } & \multicolumn{2}{c}{Average}   \\
& & NoC@85   & NoC@90 & NoC@85 & NoC@90   & NoC@85   & NoC@90 & NoC@85  & NoC@90 & NoC@85 & NoC@90  \\ 
\hline
DeepLab-LargeFOV \cite{chen2017deeplab} & $^*$ RIS-Net \cite{liew2017regional}  \ssmall{\textcolor{gray}{\textit{('17)}}} & - & 5.00 & - &  6.03 & - & - & - & -  &- & - \\ \hline
CAN \cite{YuKoltun2016} & LD \cite{li2018interactive} \ssmall{\textcolor{gray}{\textit{('18)}}} & 3.20 & 4.79 & - & - & 7.41 & 10.78 & 5.95 & 9.57 &- & - \\ \hline
 \multicolumn{1}{c|}{\multirow{2}{*}{FCN \cite{long2015fully}}} &  $^*$DOS \cite{xu2016deep} \ssmall{\textcolor{gray}{\textit{('16)}}}& 5.08 & 6.08 & - & - & 9.22 & 12.80 & 9.03 & 12.58 &- & - \\
 &  $^*$CMG \cite{majumder2019content} \ssmall{\textcolor{gray}{\textit{('19)}}} & - & 3.58 & - & 5.60 & - & - & - & - &- & - \\ \hline
DenseNet \cite{huang2017densely} & BRS \cite{jang2019interactive} \ssmall{\textcolor{gray}{\textit{('19)}}} & 2.60 & 3.60 & - & 5.08 & 6.59 & 9.78 & 5.58 & 8.24 &- & 6.68 \\ \hline
Xception-65 \cite{chollet2017xception}& $^*$CA 
\cite{kontogianni2020continuous}   \ssmall{\textcolor{gray}{\textit{('20)}}} & - & 3.07 & - & 4.94 & - & - & 5.16 & - &- & - \\ \hline
 \multicolumn{1}{c|}{\multirow{3}{*}{SegFormerB0-S2 \cite{xie2021segformer,chen2022focalclick}}} &
RITM \cite{sofiiuk2021reviving} \ssmall{\textcolor{gray}{\textit{('21)}}} & \underline{1.62} & \underline{1.82} & \textbf{1.84} & \underline{2.92} & \underline{4.26} & \underline{6.38} & \underline{4.65} & \underline{6.13} & \underline{3.09} & \underline{4.31} \\
 & FocalClick \cite{chen2022focalclick} \ssmall{\textcolor{gray}{\textit{('22)}}} & 1.66 & 1.90 & - & 3.14 & 4.34 & 6.51 & 5.02 & 7.06 & - & 4.65  \\
 &  GPCIS \ssmall{\textit{(Ours)}}    &  \textbf{1.60}  &  \textbf{1.76} &   \textbf{1.84} &  \textbf{2.70} &  \textbf{4.16} &  \textbf{6.28} &  \textbf{4.45} &  \textbf{6.04} &  \textbf{3.01} &  \textbf{4.20}  \\ \hline
 \multicolumn{1}{c|}{\multirow{3}{*}{HRNet18s-S2 \cite{wang2020deep,chen2022focalclick}}} & 
RITM \cite{sofiiuk2021reviving} \ssmall{\textcolor{gray}{\textit{('21)}}} & 2.00 & 2.24 & \underline{2.13} & 3.19 & \underline{4.29} & \underline{6.36} & \underline{4.89} & 6.54 & \underline{3.33} & 4.58 \\
 & FocalClick \cite{chen2022focalclick} \ssmall{\textcolor{gray}{\textit{('22)}}}& \underline{1.86} & \underline{2.06} & - & \underline{3.14} & 4.30 & 6.52 & 4.92 & \underline{6.48} &-& \underline{4.55}  \\
 &  GPCIS \ssmall{\textit{(Ours)}}    &  \textbf{1.74} &   \textbf{1.94} &   \textbf{1.83} &  \textbf{2.65} &  \textbf{4.28} &  \textbf{6.25} &  \textbf{4.62} &  \textbf{6.16}     &   \textbf{3.12} &  \textbf{4.25}  \\ \hline
 \multicolumn{1}{c|}{\multirow{7}{*}{ResNet50 \cite{he2016deep}}} 
  & $^*$FCANet\cite{lin2020interactive} \ssmall{\textcolor{gray}{\textit{('20)}}}   & 2.18 & 2.62 & - & 4.66 & - & - & 5.54 & 8.83 & -& -  \\
  & f-BRS-B \cite{sofiiuk2020f} \ssmall{\textcolor{gray}{\textit{('20)}}}   & 2.20 & 2.64 & 2.17 & 4.22 & 4.55 & 7.45 & 5.44 & 7.81 & 3.59 & 5.53  \\
 & CDNet \cite{chen2021conditional} \ssmall{\textcolor{gray}{\textit{('21)}}} & 2.22 & 2.64 & - & 3.69 & 4.37 & 7.87 & 5.17 & 6.66 &-& 5.22  \\
& RITM \cite{sofiiuk2021reviving} \ssmall{\textcolor{gray}{\textit{('21)}}} & 2.16 & 2.30 & 1.90 & 2.95 & 3.97 & 5.92 & \underline{4.56} & 6.05 &3.15& 4.31  \\
& FocusCut \cite{lin2022focuscut} \ssmall{\textcolor{gray}{\textit{('22)}}}  & \textbf{1.60} & \textbf{1.78} & \underline{1.86} & 3.44 & \textbf{3.62}  & \textbf{5.66} & 5.00 & 6.38 & \underline{3.02}& 4.32  \\
& FocalClick \cite{chen2022focalclick} \ssmall{\textcolor{gray}{\textit{('22)}}}  & 1.92 & 2.14 & 1.87 & \underline{2.86} & 3.84 & 5.82 & 4.61 & \underline{6.01} &3.06& \underline{4.21}   \\
 &   GPCIS \ssmall{\textit{(Ours)}}  &  \underline{1.64} &  \underline{1.82} &   \textbf{1.60} &  \textbf{2.60} &  \underline{3.80} &  \underline{5.71} &   \textbf{4.37} &  \textbf{5.89}  &   \textbf{2.85} &   \textbf{4.00} \\ \bottomrule
\end{tabular}}
\vspace{-1mm}
\end{table*}

\begin{table*}[t] 
\footnotesize
\vspace{-1mm}
\caption{Quantitative evaluation on different metrics, and comparisons on parameters and inference time. Here the backbone segmentor is ResNet50, and Second Per Click (SPC) is averagely computed over DAVIS with the testing image size of 384×384 on an NVIDIA V100 GPU. Lower {NoC$_{100}$@90}, {NoF$_{100}$@90}, NoIC, \#Params, SPC and higher IoU\&1, IoU\&5 indicate better performance.
\vspace{-3mm}}
\label{tab:comp}
\centering
\renewcommand{\arraystretch}{0.94}
\setlength{\tabcolsep}{0.94mm}{
\begin{tabular}{@{}l|cccccccccc|cc@{}}
\toprule \multicolumn{1}{l|}{\multirow{2}{*}{Method}}  & \multicolumn{5}{c|}{\multirow{1}{*}{Berkeley \cite{mcguinness2010comparative}}} & \multicolumn{5}{c|}{\multirow{1}{*}{DAVIS \cite{perazzi2016benchmark}}} & \multirow{2}{*}{{\#Params (MB)}} & \multirow{2}{*}{{SPC (ms)}} \\ 
 & { NoC$_{100}$@90} & {NoF$_{100}$@90} & {IoU\&1} & {IoU\&5} & \multicolumn{1}{c|}{{NoIC}} & { NoC$_{100}$@90} & {NoF$_{100}$@90} & {IoU\&1} & {IoU\&5} & {NoIC}{} & & \\ \hline
 f-BRS-B \cite{sofiiuk2020f}  & 6.21 & 2  & 77.06\% & 85.00\% & \multicolumn{1}{c|}{1} & 22.62 & 57 & 70.97\% & 83.87\% & \textbf{0} & \underline{39.44} & 116.53 \\ 
  CDNet \cite{chen2021conditional}  & - & - & - & - & \multicolumn{1}{c|}{-} & 18.59 & 48 & - & - & - & 39.90 & 57.76 \\ 
  RITM \cite{sofiiuk2021reviving} & \underline{3.75} & \textbf{1} & 76.88\% & 94.66\% & \multicolumn{1}{c|}{2} & 18.09 & 51 & \underline{72.89\%} & \underline{89.14\%} & 74 & {39.48} & \textbf{34.24} \\ 
FocusCut \cite{lin2022focuscut}  & 4.63 & \textbf{1} & \underline{78.89\%} & 92.89\% & \multicolumn{1}{c|}{1} & 19.00 & \underline{45} & 72.71\% & 87.58\% & 6 & 40.36 & 950.68 \\ 
FocalClick \cite{chen2022focalclick}  & 4.46 & 2 & 75.59\% & \underline{94.90\%} & \multicolumn{1}{c|}{\textbf{0}} & \underline{17.74} & 49 & 70.76\% & 88.90\% & 42 & 39.50 & 41.80 \\ 
GPCIS {\textit{(Ours)}}   & \textbf{3.36} & \textbf{1} & \textbf{79.43\%} & \textbf{95.11\%} & \multicolumn{1}{c|}{\textbf{0}} & \textbf{17.03} & \textbf{44} & \textbf{75.67\%} & \textbf{89.60\%} & \underline{2} & \textbf{39.39} & \underline{38.82}  \\ 
 \bottomrule
\end{tabular}}
\vspace{-4mm}
\end{table*}

\subsection{Model Verification}\label{sec:verf}
\noindent{\textbf{Decoupled GP posterior.}}
We firstly execute a model verification experiment to present the working mechanism underlying the decoupled GP posterior sampling framework Eq.~\eqref{eqn:inference}. From
Fig.~\ref{fig:vis}, we can clearly observe that the probability maps output by the weight-space prior term can provide rough segmentation results of the target objects. This is mainly attributed to the proposed weight space DKL strategy which can flexibly learn the prior knowledge for the IS task from the training dataset. Besides, as presented, the function-space update term compensates the prior term by utilizing relations of pixels and assigning a larger probability to unclicked pixels semantically similar to the clicks. Then it helps achieve better predictions of unclicked points by propagating the information of the clicks, such as the regions far from the click on the tiger and the long stick. Attributed to the mutual promotion of the weight-space prior and function-space update, our method obtains accurate segmentation results, approaching the ground truth (GT) masks. The results finely comply with the analysis in \textit{\textbf{\myhyperlink{remark1}{Remark 1}}} \myhyperlink{RR1}{\dcircle{1}} and validate the rationality of our proposed method.

\vspace{1mm}
\noindent{\textbf{Accuracy at clicked points.}} 
Based on the backbone ResNet50 and the DAVIS dataset, we utilize the NoIC metric to evaluate the prediction accuracy at clicks of our proposed GPCIS under different $\epsilon^2$ during testing. From Table \ref{tab:sigma} where $\epsilon^2$ is set to 0.01 during training, we can easily observe that as $\epsilon^2$ gradually gets smaller during testing, NoIC almost shows a clear downward trend, which supports the claim in \textit{\textbf{\myhyperlink{remark1}{Remark 1}}} \myhyperlink{RR2}{\dcircle{2}} that our proposed GPCIS can achieve accurate predictions at clicks with small enough $\epsilon^2$. Hence, in the following experiments, we reasonably adopt a larger $\epsilon^2$ as $10^{-2}$ for training stability and a smaller $\epsilon^2$ as $10^{-7}$ during testing for more accurate predictions at clicks.

\vspace{-1mm}
\subsection{Performance Evaluation}
In this section, based on the four datasets, \emph{i.e.}, GrabCut, Berkeley, SBD, and DAVIS, we comprehensively validate the effectiveness of our proposed method by comparing it with a series of IS methods~\cite{liew2017regional,li2018interactive,xu2016deep,majumder2019content,jang2019interactive,kontogianni2020continuous,chen2022focalclick,lin2020interactive,sofiiuk2020f,chen2021conditional,lin2022focuscut}. For fair comparisons with the current state-of-the-art (SOTA) methods~\cite{chen2022focalclick,lin2020interactive,sofiiuk2020f,chen2021conditional,lin2022focuscut}, we separately implement our proposed GPCIS with the backbone segmentor SegFormerB0-S2 and HRNet18s-S2 adopted by \cite{chen2022focalclick}, and with ResNet50 widely adopted by \cite{lin2020interactive,sofiiuk2020f,chen2021conditional,lin2022focuscut}. Note that our proposed method is orthogonal to most of the competitors and yet we do not adopt their exclusive designs, such as cropping click-centered patches with adaptive scopes in FocusCut~\cite{lin2022focuscut}, and local refinement and progressive merge in FocalClick \cite{chen2022focalclick}. RITM~\cite{sofiiuk2021reviving} is also reimplemented as our baseline under the same experimental settings.~\footnote{More experimental results are provided in \textit{SM}.}

\begin{figure*}[t]
  \centering
  \vspace{-2mm}
\includegraphics[width=0.98\linewidth]{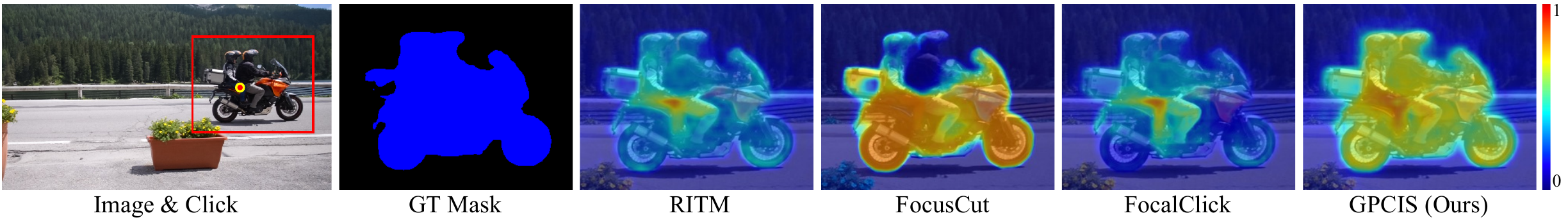}
  \vspace{-0.3cm}
  \caption{{Visualization comparisons on the probability maps output by different competing methods.}}
  \label{fig:compare2}
  \vspace{-0.2cm}
\end{figure*}

\begin{figure}[t]
  \centering
  \vspace{-0.3cm}
\includegraphics[width=0.95\linewidth]{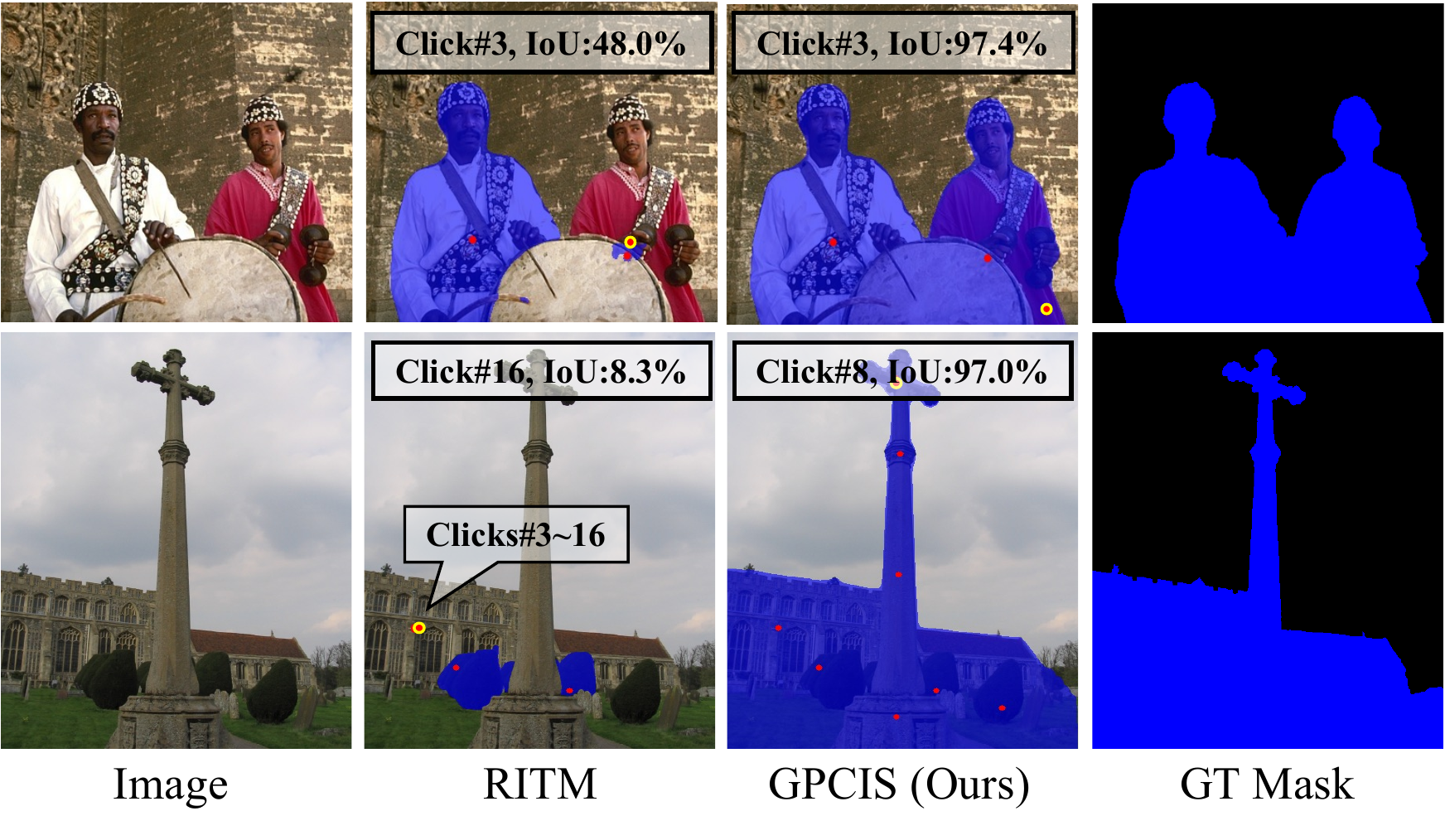}
  \vspace{-0.2cm}
  \caption{{Exemplar segmentation results of RITM \cite{sofiiuk2021reviving} and GPCIS.}}
  \label{fig:compare}
  \vspace{-0.4cm}
\end{figure}

\vspace{1mm}
\noindent \textbf{Quantitative evaluation.} Table~\ref{tab:exp} lists the NoC@85 and NoC@90 of all the comparing methods on the four different datasets. We can clearly find that the proposed GPCIS consistently achieves the lowest average NoC@85 and NoC@90 under three different backbone segmentors, which substantiates its promising effectiveness and good generality. Note that although our method does not introduce the extra processing strategies contained in the SOTA method FocusCut, \emph{e.g.}, cropping click-centered patches with adaptive scopes, it can still obtain the superior (Berkeley \& DAVIS) or at least comparable (GrabCut \& SBD) performance to FocusCut.

For comprehensive comparisons, we provide more quantitative results on different metrics as well as the number of network parameters and inference efficiency. As listed in Table~\ref{tab:comp}, the proposed GPCIS consistently outperforms other competing methods on NoC$_{100}$@90, NoF$_{100}$@90, IoU\&1, IoU\&5, and the model size, where NoC$_{100}$@90 and NoF$_{100}$@90 represent the numbers of clicks and failures to get 90\% IoU within 100 clicks, respectively. For NoIC and SPC, it still performs competing and is comparable to the first rank. From Table~\ref{tab:exp} and Table~\ref{tab:comp}, we can easily conclude that compared to other comparing methods, our proposed GPCIS shows better generality and it has the capability to efficiently attain higher segmentation accuracy with fewer clicks and fewer failure cases. This indicates that our method has good potential for practical IS. Note that compared to the baseline RITM, our inference speed is slightly slower due to the proposed GP posterior inference procedure. However, this cost is acceptable or even negligible considering the performance gains brought by our method.

\vspace{2mm}
\noindent \textbf{Qualitative evaluation.} 
Fig. \ref{fig:compare2} presents the visualization comparisons on the output probability maps of different methods. 
As seen, for RITM and FocalClick, the regions far from the click cannot be properly and fully activated and have low prediction probability. Although FocusCut confidently segments the main part of the object, it mistakenly leaves out the upper part with low prediction probability. Comparatively, our proposed GPCIS achieves better segmentation results and approaches the GT mask, which is mainly attributed to the explicit  modeling of the semantic relations between pixels. To fully substantiate the effectiveness of our proposed inference process, we also provide more visual comparisons with the baseline RITM.  From the first row in Fig.~\ref{fig:compare}, we can observe that without fully utilizing the information contained in clicks, RITM fails to finely segment the whole target object. In contrast, our method almost accomplishes the accurate segmentation of the three target instances, \ie, two persons and a drum, within three clicks. Besides, the second row shows that from the 3$^{\text{rd}}$ to the 16$^{\text{th}}$ clicks, RITM repetitively clicks in the same location because it cannot provide correct predictions at clicks. However, with good theoretical support, GPCIS alleviates this issue and obtains a 97\% IoU within 8 clicks. 

\subsection{Ablation Studies}\label{sec:abla}
Based on the backbone segmentor ResNet50, we execute an ablation study to quantitatively evaluate the effect of the modules involved in our method on the average NoC@85/90 over GrabCut, Berkeley, SBD and DAVIS. Table \ref{tab:ablation} reports the results under different settings where variant (e) is the final strategy we adopt in comparison experiments above. By comparing (a) and (e), we can easily find that the proper guidance of $\mathcal{L}_{\mathit{VI}}$ is indeed helpful for network learning. In (b), we discard the deep kernel learning mechanism in function space and fix the kernel hyperparameters as $\eta_0=1$ and $\eta_t=e^{-1}$ ($t=1,2,\ldots, d$). Similarly, in (c), we discard the deep kernel learning mechanism in weight space and set $\bm{\theta}_r\sim\mathcal{N}(0,\mathbf{I}_d)$, $\tau_r\sim U(0,2\pi)$, $\bm{\mu}_{\mathbf{w}}\sim\mathcal{N}(0,0.25\mathbf{I}_d)$, and $\sigma^2_{\mathbf{w}}=0.025$. During network training, they are not updated. As expected, without the DKL design in double space, the network flexibility is weakened, leading to degraded performance. Besides, by comparing (d) and (e), it shows that the concatenation of input image $\mathcal{I}$ with deep features $\mathbf{X}$ shown in Fig.~\ref{fig:main} can further boost the information propagation across pixels and bring better segmentation performance.

\begin{table}[]
\footnotesize
\vspace{-0.3cm}
\caption{Ablation study on our specific designs, including $\mathcal{L}_{VI}$, double space DKL, and whether to concatenate features with $\mathcal{I}$. \vspace{-3mm}}
\label{tab:ablation}
\centering
\renewcommand{\arraystretch}{0.99}
\setlength{\tabcolsep}{0.5mm}{
\begin{tabular}{@{}c|cccc|cc@{}} 
\toprule 
Variants & $\mathcal{L}_{\mathit{VI}}$ & DKL-F & DKL-W & Concat $\mathcal{I}$  & Avg. NoC@85 & Avg. NoC@90 \\ \hline
(a) & \xmark  &   \cmark & \cmark &  \cmark & 2.98 & 4.07     \\
(b)& \cmark  & \xmark  &  \cmark  & \cmark  & 3.00 &  4.16     \\
(c)& \cmark & \cmark & \xmark  & \cmark & 3.10 & 4.34 \\
(d)& \cmark &   \cmark  &  \cmark& \xmark & 2.96 & 4.10  \\ \hline
(e)& \cmark & \cmark  & \cmark  & \cmark &  \textbf{2.85} & \textbf{4.00}  \\ \bottomrule
\end{tabular} } \vspace{-5mm}
\end{table}

\vspace{-1mm}
\section{Conclusion}
In this paper, for the interactive segmentation task, we have dived into a new perspective and regarded it as a pixel-wise binary classification problem on each input image. Based on such understanding, we have formulated the task as a Gaussian process classification model. To solve this model, we have proposed to variationally approximate the GP posterior in a data-driven manner, along with a decoupled sampling strategy with linear complexity. Correspondingly, we have constructed an efficient and flexible GP classification framework integrated with double space deep kernel learning, called GPCIS, which has clear working mechanism. Based on several benchmark datasets and different backbone segmentors, we have conducted comprehensive experiments as well as model verification, which fully substantiated the superiority of our proposed GPCIS as well as its rational theoretical support for correct predictions at clicks. 
With high efficiency and fine generality, the proposed GPCIS should be a potential driver for the interactive segmentation field.


{\small
\bibliographystyle{ieee_fullname}
\bibliography{egbib}
}

\end{document}